\documentclass[10pt,twocolumn,letterpaper]{article}

\usepackage[pagenumbers]{cvpr} 

\usepackage{amsmath}
\usepackage{mathtools}
\usepackage{pifont}
\usepackage{multicol}
\usepackage{multirow}
\usepackage{makecell}
\usepackage{colortbl}
\usepackage{float}

\definecolor{mygray}{gray}{0.94}
\definecolor{lightgray}{gray}{0.75}
\definecolor{first}{rgb}{1,0.7,0.7}
\definecolor{second}{rgb}{1,0.85,0.7}
\definecolor{third}{rgb}{1,1,0.8}

\newcommand{\name}{MoVieS\xspace}
\newcommand{\no}{{\textcolor[rgb]{0.8,0,0}{\ding{55}}}}
\newcommand{\ye}{{\textcolor[rgb]{0,0.8,0}{\ding{52}}}}
\newcommand{\na}{{\textcolor[rgb]{0,0,0.8}{\ding{52}\rotatebox[origin=c]{-9.2}{\kern-0.7em\ding{55}}}}}
\newcommand\nnfootnote[1]{
  \begin{NoHyper}
      \renewcommand\thefootnote{}\footnote{#1}
      \addtocounter{footnote}{-1}
  \end{NoHyper}
}

\definecolor{cvprblue}{rgb}{0.21,0.49,0.74}
\usepackage[pagebackref,breaklinks,colorlinks,allcolors=cvprblue]{hyperref}

\def\paperID{32421} 
\def\confName{CVPR}
\def\confYear{2026}

\title{MoVieS: Motion-Aware 4D Dynamic View Synthesis in One Second}

\author{
Chenguo Lin$^{1*}$, Yuchen Lin$^{1,3*}$, Panwang Pan$^{2\dagger}$, \\ Yifan Yu$^2$, Tao Hu$^2$, Honglei Yan$^2$, Katerina Fragkiadaki$^3$, Yadong Mu$^{1\ddagger}$ \\
$^1$Peking University, $^2$ByteDance, $^3$Carnegie Mellon University \\
\textbf{\url{https://chenguolin.github.io/projects/MoVieS}}
}

\begin{document}

\maketitle

\nnfootnote{$^*$: Equal contribution; $\dagger$: Project lead; $\ddagger$: Corresponding author.}

\begin{abstract}
    We present \textbf{\name}, a \underline{\textbf{Mo}}tion-aware \underline{\textbf{Vie}}w \underline{\textbf{S}}ynthesis model that reconstructs 4D dynamic scenes from monocular videos in one second.
It represents dynamic 3D scenes with pixel-aligned Gaussian primitives and explicitly supervises their time-varying motions.
This allows, for the first time, the unified modeling of appearance, geometry and motion from monocular videos, and enables reconstruction, view synthesis and 3D point tracking within a single learning-based framework.
By bridging view synthesis with geometry reconstruction, \name enables large-scale training on diverse datasets with minimal dependence on task-specific supervision.
As a result, it also naturally supports a wide range of zero-shot applications, such as scene flow estimation and moving object segmentation.
Extensive experiments validate the effectiveness and efficiency of \name across multiple tasks, achieving competitive performance while offering several orders of magnitude speedups.

\end{abstract}

\vspace{-0.5cm}
\section{Introduction}\label{sec:intro}
Humans and animals perceive a continuous stream of observations from a dynamic 3D world, and effortlessly interpret its underlying geometry and motion.
Replicating this capability is essential for any embodied agent that must understand and act in the physical world.

Recent advances have made great strides in individual 3D tasks, such as monocular depth estimation~\citep{ranftl2020towards,ke2024repurposing,yang2024depth,bochkovskii2024depth,chen2025video}, 3D scene reconstruction~\citep{schonberger2016structure,teed2021droid,wang2024dust3r,zhang2024monst3r,wang2025vggt}, novel view synthesis~\citep{mildenhall2020nerf,fridovich2023k,kerbl20233d,wu20244d,luiten2024dynamic} and point tracking~\citep{harley2022particle,doersch2023tapir,wang2023tracking,karaev2024cotracker,xiao2024spatialtracker}.
However, besides treating each task in isolation, most existing view synthesis and reconstruction studies focus on static scenes and require costly per-scene optimization without learning prior knowledge.
Real-world environments are inherently dynamic and diverse, and all the aforementioned 3D scene understanding tasks could share common underlying principles.

\begin{figure*}[t]
    \begin{center}
    \includegraphics[width=0.95\textwidth]{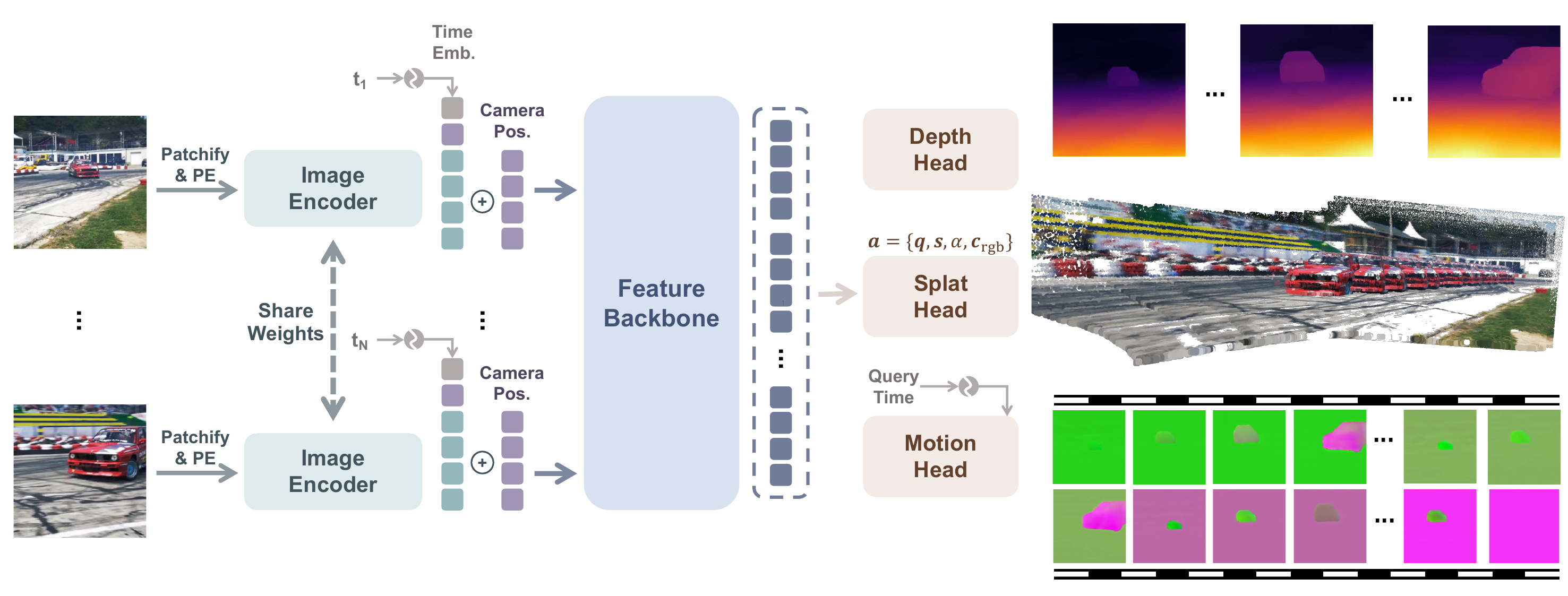}
    \end{center}
    \vspace{-0.6cm}
    \caption{\textbf{Overview}. \name consists of a shared image encoder, an attention-based feature backbone (Sec.~\ref{subsubsec:backbone}), and three heads (Sec.~\ref{subsubsec:heads}) to jointly model appearance, geometry and motion. Motion head is time-conditioned to model dynamic content with respect to several query timestamps. Normalized XYZ values in the 3D space of motion maps are treated as RGB channels for visualization. Time-varying Gaussian attributes are omitted and point clouds with color and identity Gaussian attributes are visualized here for brevity.}
    \label{fig:overview}
    \vspace{-0.4cm}
\end{figure*}

Motivated by this, we introduce \textbf{\name}, a \underline{\textbf{Mo}}tion-aware dynamic \underline{\textbf{Vie}}w \underline{\textbf{S}}ynthesis model for feed-forward 4D reconstruction of monocular videos, that jointly models scenes' appearance, geometry and motion.
\name represents 3D dynamic scenes using renderable and deformable 3D particles, termed \textbf{dynamic splatter pixels}, and utilizes a differentiable 3D Gaussian rendering framework~\citep{kerbl20233d}.
Specifically, following recent practices on feed-forward view synthesis~\citep{charatan2024pixelsplat,szymanowicz2024splatter,zhang2024gs,xu2024depthsplat,lin2025diffsplat}, each input pixel is mapped to a 3D Gaussian primitive, with its 3D location determined by predicted depth.
To model dynamics, \name regresses per-pixel motion displacements toward arbitrary query timestamps, enabling temporal tracking of each splatter pixel.
This design facilitates coherent reconstruction of both 3D geometry and appearance across camera viewpoints and temporal frames.

As shown in Figure~\ref{fig:overview}, \name is built upon a large-scale pretrained transformer backbone~\citep{oquab2024dinov2,wang2025vggt}, which encodes each video frame independently and aggregates their information via attentions~\citep{vaswani2017attention}.
The aggregated features are then processed by specialized prediction heads: 
(1) a \textbf{depth head} estimates depth for each input frame, (2) a \textbf{splatter head} predicts per-pixel 3D Gaussian~\citep{kerbl20233d} appearance attributes, such as color and opacity, for novel view rendering, (3) a \textbf{motion head} estimates the time-conditioned movements of Gaussian primitives towards a target timestamp, allowing us to track its temporal evolution.

Benefiting from the unified architecture, \name can be trained on large-scale datasets featuring both static~\citep{zhou2018stereo,wang2020tartanair,li2023matrixcity} and dynamic~\citep{mehl2023spring,cabon2020virtual} scenes, as well as point tracking datasets~\citep{zheng2023pointodyssey,karaev2023dynamicstereo,jin2024stereo4d}.
During inference, it takes a monocular video, whether depicting a static or dynamic scene, and reconstructs per-pixel 3D Gaussian primitives along with their motion attributes at any target timestamp, enabling view synthesis, depth estimation, and 3D point tracking in a single model.

Extensive experiments on diverse benchmarks~\citep{zhou2018stereo,yoon2020novel,gao2022dynamic,koppula2024tapvid} reveal that \name achieves competitive performance across a variety of 4D perception tasks, while being several orders of magnitude faster than the existing state of the art.
Furthermore, empowered by novel view synthesis as a proxy task, \name enables dense motion understanding from sparse tracking supervision.
This naturally gives rise to various applications, such as scene flow estimation and moving object segmentation, further broadening the potential of our approach.

In summary, our main contributions include:
\vspace{0.1cm}
\begin{itemize}
    \item We introduce \name, a novel feed-forward framework that jointly models appearance, geometry and motion for 4D scene perception from monocular videos.
    \item Dynamic splatter pixels are proposed to represent dynamic 3D scenes as renderable deforming 3D particles, bridging novel view synthesis and dynamic geometry reconstruction.
    \item \name delivers strong performance and orders of magnitude speedups for 4D reconstruction, and naturally enables a wide range of applications in a zero-shot manner.
\end{itemize}

\vspace{-0.2cm}
\section{Related Work}\label{sec:related}

\begin{figure*}[t]
    \begin{center}
    \includegraphics[width=0.95\textwidth]{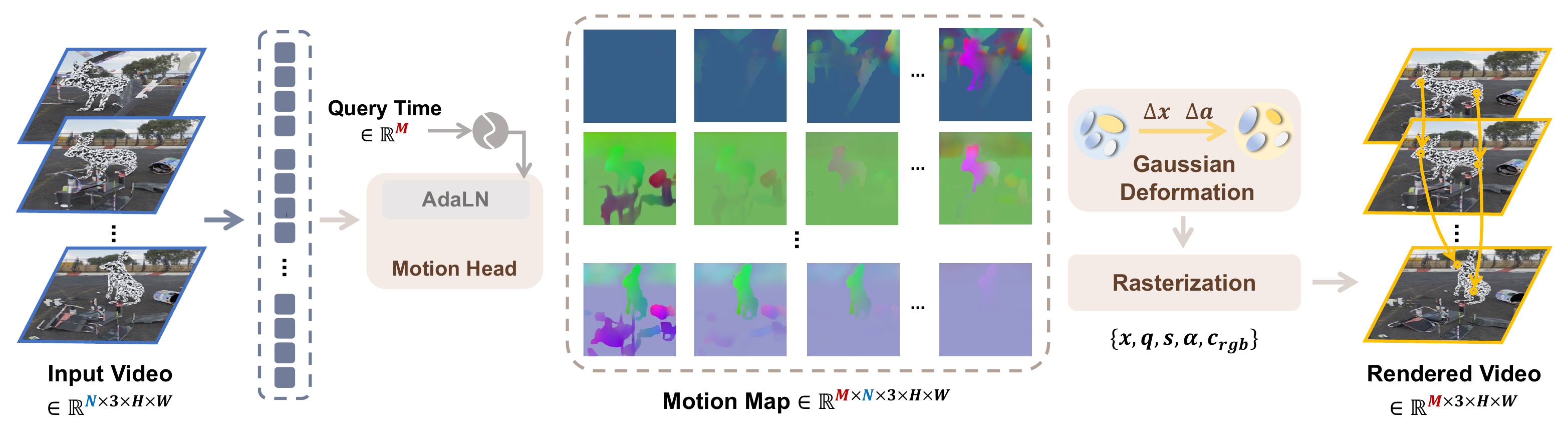}
    \end{center}
    \vspace{-0.6cm}
    \caption{\textbf{Motion Head}.
    Given $M$ query timesteps, the proposed motion head is conditioned via adaptive layer normalization (AdaLN) and predicts 3D displacements for each input pixel. After rasterization using the $M$ corresponding query-time cameras, output images in shape $M\times 3\times H\times W$ are rendered for supervision. Gaussian attribute deformation $\Delta\mathbf{a}$ is omitted for brevity.}
    \label{fig:motion_head}
    \vspace{-0.4cm}
\end{figure*}

\vspace{-0.1cm}
\subsection{Feed-forward 3D Reconstruction}
Traditional 3D reconstruction relies on dense multi-view supervision and per-scene optimization~\citep{hartley2003multiple,mur2015orb,schonberger2016structure,mildenhall2020nerf}.
Recent learning-based methods leverage large-scale priors to enable feed-forward prediction of depth, pose, and other 3D properties~\citep{wang2024vggsfm,ranftl2020towards,piccinelli2024unidepth,hu2024metric3d}.
With differentiable rendering techniques such as 3D Gaussian Splatting (3DGS)~\citep{kerbl20233d}, some works further regress pixel-wise Gaussian attributes for novel view synthesis~\citep{charatan2024pixelsplat,chen2024mvsplat,zhang2024gs}.
DUSt3R~\citep{wang2024dust3r} pioneered the direct regression of pixel-aligned pointmaps in a canonical space from image pairs.
Its extensions~\citep{leroy2024grounding,wang2025d,liu2024slam3r,smart2024splatt3r,zhang2025flare} generalize this framework with multi-view, streaming, and 3DGS integration.
VGGT~\citep{wang2025vggt} unifies these advances with a strong image encoder~\citep{oquab2024dinov2} and task-specific heads.
However, all of these works are limited to static scenes.
In contrast, in this work, we extend feed-forward 3DGS reconstruction to dynamic scenarios with moving objects.

\vspace{-0.1cm}
\subsection{Dynamic Reconstruction and View Synthesis}
Compared to static reconstruction, dynamic reconstruction remains underexplored.
Extensions of DUSt3R~\citep{chen2025easi3r,sucar2025dynamic,jin2024stereo4d} handle dynamic reconstruction in a plug-and-play manner~\citep{chen2025easi3r} or leveraging foundation models~\citep{zhang2024monst3r,lu2024align3r,yao2025uni4d} with monocular depth~\citep{bochkovskii2024depth,piccinelli2025unidepthv2}, optical flow~\citep{teed2020raft}, and point tracking~\citep{karaev2024cotracker}, but remain limited to two-frame inputs and output only sparse point clouds.
CUT3R~\citep{wang2025continuous} extends to video inputs via recurrent updates, while diffusion-based methods~\citep{xu2025geometrycrafter,jiang2025geo4d} treat reconstruction as conditional generation, though at the cost of multiple passes per sequence.

Instead of geometric structure, dynamic novel view synthesis focuses on rendering quality and view-dependent effects.
NeRF-based methods~\citep{mildenhall2020nerf,fridovich2023k,cao2023hexplane} reconstruct dynamic scenes from multi-view~\citep{bansal20204d,wang2022fourier,li2022neural} or monocular videos~\citep{pumarola2021d,park2021nerfies,li2023dynibar,zhao2024pgdvs}, while 3DGS~\citep{kerbl20233d} extends this with 4D primitives~\citep{yang2024real,li2024spacetime} or deformable fields~\citep{wu20244d,luiten2024dynamic,yang2024deformable,lin2024gaussian,liu2025modgs}.
Recent works further express Gaussian motion explicitly~\citep{wang2024shape,sun2024splatter,stearns2024dynamic,lei2025mosca,lin2025omniphysgs}, but require scratch training, iterative optimization, and external supervision from point tracking or optical flow estimation~\citep{doersch2023tapir,teed2020raft}.

For feed-forward 4D view synthesis, STORM~\citep{yang2025storm} is designed for outdoor driving scenes from multi-view videos.
BTimer~\citep{liang2024feed} and NutWorld~\citep{shen2025seeing} estimate 3DGS attributes from monocular inputs.
However, BTimer predicts independent pixel-wise 3D Gaussians per timestamp without modeling their frame-wise relations and needs an extra enhancer module for smooth intermediate frames.
NutWorld models Gaussian motion but lacks explicit supervision, relying heavily on pretrained depth~\citep{chen2025video} and flow~\citep{xu2023unifying} estimation models, and uses an orthographic camera, which further leads to projection distortions.

\vspace{-0.2cm}
\section{Method}\label{sec:method}
Following previous studies on monocular 4D reconstruction~\citep{zhao2024pgdvs,wang2024shape,liu2025modgs,lei2025mosca,liang2024feed,yang2025storm}, given a \textit{posed} video $\mathcal{V}\coloneqq\{\mathbf{I}_i, \mathbf{P}_i, \mathbf{K}_i, t_i\}_{i=1}^N$ with frames $\mathbf{I}_i\in\mathbb{R}^{3\times H\times W}$, camera poses $\mathbf{P}_i\in\mathbb{SE}(3)$, intrinsics $\mathbf{K}_i\in\mathbb{R}^{3\times 3}$ and timestamps $t_i\in[0,1]$, we aim to train a generic model that jointly reconstructs its appearance, geometry, and motion.
We leave the handling of unposed videos to future work.

\vspace{-0.1cm}
\subsection{Dynamic Splatter Pixel}\label{subsec:repre}
To model 3D scenes with moving contents, we propose a novel representation, namely \textbf{dynamic splatter pixel}, which decomposes dynamic scenes into a set of static Gaussian primitives and their corresponding deformation fields.
Given an input video $\mathcal{V}$, each pixel in the $i$-th frame $\mathbf{I}_i$ is associated with a splatter pixel $\mathbf{g}_i$~\citep{charatan2024pixelsplat,szymanowicz2024splatter,zhang2024gs} in a \textit{shared} canonical space of the first frame camera's coordinate system.
Each $\mathbf{g}\coloneqq\{\mathbf{x},\mathbf{a}\}$ is parameterized by its position $\mathbf{x} \in \mathbb{R}^3$ in the canonical space and other rendering attributes $\mathbf{a}\in\mathbb{R}^{11}$, including rotation quaternion $\mathbf{q} \in \mathbb{R}^4$, scale $\mathbf{s} \in \mathbb{R}^3$, opacity $\alpha \in \mathbb{R}$, and color $\mathbf{c}_{\text{rgb}} \in \mathbb{R}^3$~\citep{kerbl20233d}.
Considering splatter pixels were originally designed for static scenes, we decouple motion from geometric structure to adapt them for dynamic scenes.
An additional time-dependent \textbf{deformation field} is introduced, in which each splatter pixel $\mathbf{g}$ is associated with $\mathbf{m}(t) \coloneqq\{\Delta\mathbf{x}(t), \Delta\mathbf{a}(t)\}$.
$\Delta\mathbf{x}(t) \in \mathbb{R}^3$ is the motion vector of a splatter pixel at time $t$ with respect to the \textit{canonical} 3D space, and $\Delta\mathbf{a}(t)$ is the change of the corresponding attributes of the splatter pixel at time $t$.
Therefore, the splatter pixel $\mathbf{g}$ is deformed to time $t$ as:
\begin{equation}
    \mathbf{x} \gets \mathbf{x} + \Delta\mathbf{x}(t), \quad \mathbf{a} \gets \mathbf{a} + \Delta\mathbf{a}(t).
\end{equation}
By combining static splatter pixels and their deformation fields, we establish the correspondence between each Gaussian primitive and its temporal dynamics, thereby enabling dynamic scene modeling and dense motion estimation.

\vspace{-0.15cm}
\subsection{Unify Appearance, Geometry and Motion}\label{subsec:model}

\subsubsection{Feature Backbone}\label{subsubsec:backbone}
Given a posed video $\mathcal{V}=\{\mathbf{I}_i, \mathbf{P}_i, \mathbf{K}_i,t_i\}_{i=1}^N$, we first patchify each input image $\mathbf{I}_i$ and use a pretrained image encoder~\citep{oquab2024dinov2} to extract their features as shown in Figure~\ref{fig:overview}.
To effectively incorporate camera information, we adopt two complementary strategies to embed the camera parameters into image features:
(1) \textbf{Pl\"ucker embedding}: camera pose $\mathbf{P}_i$ and intrinsics $\mathbf{K}_i$ are transformed into pixel-aligned Pl\"ucker embeddings~\citep{sitzmann2021light}, which are then downsampled and fused with image features via spatial-wise addition;
(2) \textbf{Camera token}: $\mathbf{P}_i$ and $\mathbf{K}_i$ are passed through a linear layer and encoded to a camera token, which is appended to the sequence of image tokens.
Ablation study on these two camera injection manners is validated in Sec.~\ref{subsec:ablation}.

To inform the model that input images originate from a temporally ordered video, we additionally encode the timestamp $t_i\in[0,1]$ of input frames by sinusoidal positional encoding~\citep{mildenhall2020nerf} to produce a \textbf{timestamp token}, which is then concatenated with the aforementioned image and camera tokens as shown in Figure~\ref{fig:overview}.
After tokenizing input images, camera parameters, and timestamps, we apply the geometrically pretrained attention blocks~\citep{vaswani2017attention} from VGGT~\citep{wang2025vggt} to enable interactions among image tokens across video frames.
This produces a set of shared feature tokens enriched with inter-frame context as well as camera and temporal information, which are then used for predicting various properties of dynamic scenes.
The effect of VGGT initialization is discussed in Sec.~\ref{subsec:ablation}.

\subsubsection{Prediction Heads}\label{subsubsec:heads}
Tokens from the feature backbone are fed into three parallel prediction heads that estimate the appearance, geometry and motion of a dynamic scene respectively, as shown in Figure~\ref{fig:overview}.
Each head adopts a DPT-style architecture~\citep{ranftl2021vision} to convert aggregated tokens into dense predictions matching the input resolution and product dynamic splatter pixels.

\vspace{-0.2cm}
\paragraph{Depth and Splatter Head}
Different from previous feed-forward 3DGS reconstruction methods~\citep{charatan2024pixelsplat,chen2024mvsplat,zhang2024gs} that use a single head to predict all splatter pixel attributes, we adopt a decoupled design to better leverage geometric priors from the pretrained VGGT~\citep{wang2025vggt}.
Specifically, a dedicated \textbf{depth head}, initialized from VGGT, is used for geometry prediction to provide spatial grounding for splatter pixel construction, while another DPT as \textbf{splatter head} is trained from scratch for appearance rendering.
We further incorporate a direct RGB shortcut~\citep{ye2024no} from the input image to the final convolution layer of the splatter head to preserve high-frequency details and enhance color fidelity.

\vspace{-0.2cm}
\paragraph{Motion Head}
To capture scene dynamics, a novel \textbf{motion head} as shown in Figure~\ref{fig:motion_head} is introduced to predict dense deformation $\mathbf{m}(t)$ for each dynamic splatter pixel at any target moment.
Temporal variation is enabled by injecting the sinusoidally encoded query time $t_q$ into the aggregated tokens via adaptive layer normalization~\citep{xu2019understanding} before applying DPT convolutions.
For each input frame $\mathbf{I}_i$ at time $t_i$, the motion head predicts its 3D movements $\Delta\mathbf{x}$ and Gaussian attribute deformation $\Delta\mathbf{a}$ toward $t_q$ in a \textit{shared} world coordinate system.
To visualize motion maps, the XYZ coordinates in 3D space are jointly normalized by their shared minimum and maximum values to the range $[0, 1]$, and then mapped to RGB channels for visualization.

\begin{table}[t]
    \centering
    \caption{\textbf{Training Datasets}. Eight datasets from diverse sources are utilized to train \name at scale.}
    \label{tab:datasets}
    \renewcommand\arraystretch{1.1}
    \resizebox{0.475\textwidth}{!}{

    \begin{tabular}{lccccr}
        \toprule

        \textbf{Dataset} & \textbf{Dynamic?} & \textbf{Depth?} & \textbf{Tracking?} & \textbf{Real?} & \textbf{\#Scenes} \\

        \midrule

        RealEstate10K~\citep{zhou2018stereo} & \no & \no & \no & \ye & 70K \\
        TartanAir~\citep{wang2020tartanair} & \no & \ye & \no & \no & 0.4K \\
        MatrixCity~\citep{li2023matrixcity} & \no & \ye & \no & \no & 4.5K \\
        PointOdyssey~\citep{zheng2023pointodyssey} & \ye & \ye & \ye & \no & 0.1K \\
        DynamicReplica~\cite{karaev2023dynamicstereo} & \ye & \ye & \ye & \no & 0.5K \\
        Spring~\citep{mehl2023spring} & \ye & \ye & \no & \no & 0.03K \\
        VKITTI2~\citep{cabon2020virtual} & \ye & \ye & \no & \no & 0.1K \\
        Stereo4D~\citep{jin2024stereo4d} & \ye & \ye & \ye & \ye & 98K \\

        \bottomrule
    \end{tabular}

    }
\end{table}

\begin{figure*}[t]
    \begin{center}
    \includegraphics[width=0.95\textwidth]{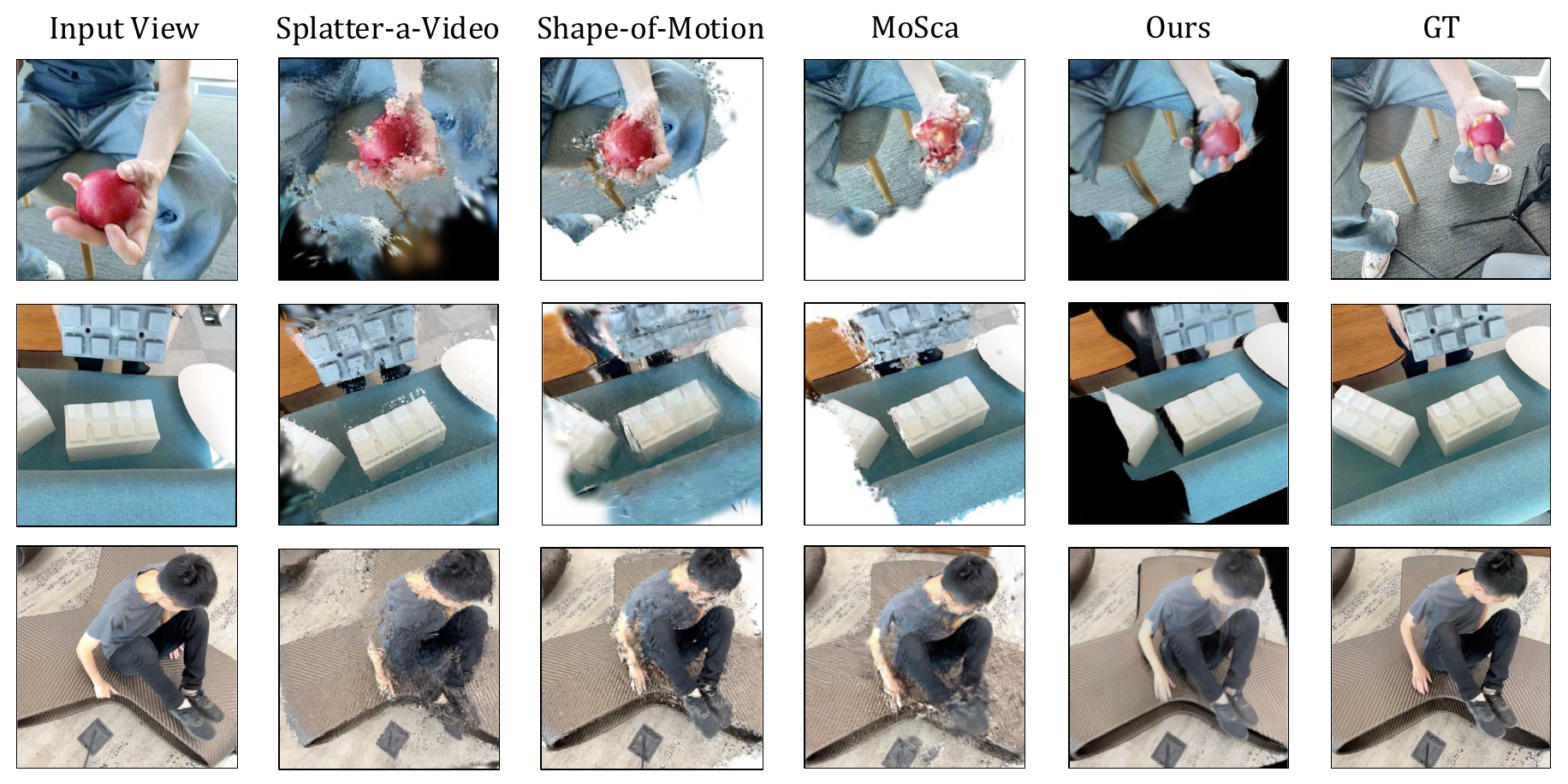}
    \end{center}
    \vspace{-0.4cm}
    \caption{\textbf{Novel View Synthesis for Dynamic Scenes}. Given a monocular video, we compare synthesized novel views of different methods. Invisible regions are rendered as black or white, depending on the implementation. More results are in the supplementary material.}
    \label{fig:nvs_vis}
    \vspace{-0.2cm}
\end{figure*}

\subsection{Training}\label{subsec:training}

\subsubsection{Dataset Curation}
An ideal dataset for dynamic reconstruction would include synchronized multi-view videos with dense depth and point tracking annotations.
However, such data is infeasible to capture and annotate at scale in practice.
Instead, we leverage a diverse set of open-source datasets~\citep{zhou2018stereo,wang2020tartanair,li2023matrixcity,mehl2023spring,cabon2020virtual,zheng2023pointodyssey,karaev2023dynamicstereo,jin2024stereo4d}, each providing \textit{complementary} supervision, as shown in Table~\ref{tab:datasets}.
With the flexible model design, \name can be trained on these heterogeneous sources by aligning objectives to their respective annotations.
For datasets without an official test split, we randomly sample 10\% of the data as the test set for generalization evaluation.
Details about data curation are in the supplementary material.

\subsubsection{Objectives}
\name is trained by a multi-task objective that combines depth, rendering and motion losses:
\begin{equation}
    \mathcal{L}\coloneqq \lambda_{\text{d}}\mathcal{L}_{\text{depth}} + \lambda_{\text{r}}\mathcal{L}_{\text{rendering}} + \lambda_{\text{m}}\mathcal{L}_{\text{motion}}.
    \label{equ:loss}
\end{equation}

\vspace{-0.4cm}
\paragraph{Depth and Rendering Losses}
Depth loss is computed as the mean squared error (MSE) between the predicted depth maps $\hat{D}_i$ and ground truth depth maps $D_i$, along with their spatial gradients, after filtering out invalid values.
\vspace{-0.1cm}
\begin{equation}
    \mathcal{L}_{\text{depth}}\coloneq \sum_{i=1}^N \|D_i-\hat{D}_i\|_2 + \|\nabla D_i - \nabla \hat{D}_i\|_1.
\end{equation}\vspace{-0.1cm}Rendering loss combines pixel-wise MSE and perceptual loss~\citep{zhang2018perceptual} between 3DGS-rendered images $\hat{I}$ under $M$ camera views and the video frames at corresponding target timestamps, which are randomly sampled during training.
\vspace{-0.1cm}
\begin{equation}
    \mathcal{L}_{\text{rendering}}\coloneq \sum_{v=1}^{M}\|I_v-\hat{I}_v\|_2 + \lambda_{\text{LPIPS}} \cdot \text{LPIPS}(I_v, \hat{I}_v).
\end{equation}\vspace{-0.1cm}Weighting terms $\lambda_\text{d}$, $\lambda_\text{r}$ and $\lambda_{\text{LPIPS}}$ are set to $1$, $1$ and $0.5$ by default respectively.

\vspace{-0.3cm}
\paragraph{Motion Loss}
Given 3D point tracking datasets, ground-truth motion $\Delta\mathbf{x}$ is defined as the 3D displacement of each tracked point between any two frames.
Since all 3D points are defined in the world coordinate and most tracked points remain static, implying that their corresponding motion vectors tend to zero.
We apply a point-wise L1 loss between the predicted and ground-truth motions to promote sparsity after filtering out points that are not visible in the input frames.
Additionally, to complement direct point-to-point alignment, a \textbf{distribution loss} is introduced that encourages the predicted motion vectors to preserve the internal relative distance structure within each frame.
The final motion loss is defined as a combination of point-wise and distribution-level supervision:
{\small\begin{align}
    \mathcal{L}_{\text{motion}}
    &\coloneqq \lambda_{\text{pt}}\mathcal{L}_{\text{pt}} + \lambda_{\text{dist}}\mathcal{L}_{\text{dist}}\label{equ:motion}\\
    &\ = \frac{1}{P}\sum_{i\in\Omega}\lambda_{\text{pt}}\|\Delta\hat{\mathbf{x}}_i-\Delta\mathbf{x}_i\|_1\\
    &\ + \frac{1}{P^2}\sum_{(i,j)\in\Omega\times\Omega}\lambda_{\text{dist}}\|\Delta\hat{\mathbf{x}}_i\cdot\Delta\hat{\mathbf{x}}_j^\top - \Delta\mathbf{x}_i\cdot\Delta\mathbf{x}_j^\top\|_1,
\end{align}}where $\Omega$ denotes the set of valid $P$ tracked points, and $\Omega \times \Omega$ is its Cartesian product.
Weighting term $\lambda_\text{m}$, $\lambda_{\text{pt}}$, and $\lambda_{\text{dist}}$ are set to $10$, $1$, and $10$ respectively.
Ablation study on the effectiveness of these two types of motion supervision is presented in Sec.~\ref{subsec:ablation}.

\vspace{-0.3cm}
\paragraph{Normalization}
Similar to VGGT~\citep{wang2025vggt}, we normalize the 3D scene scale by the average Euclidean distance from each 3D point to the origin of the canonical world coordinate system.
As a result, unlike some other reconstruction methods~\citep{wang2024dust3r,leroy2024grounding,wang2025continuous}, we do not apply additional normalization in the depth or motion loss.
We also omit confidence-aware weighting~\citep{wang2024dust3r,wang2025vggt} for simplicity and more stable training.

\vspace{-0.2cm}
\section{Experiments}\label{sec:exp}

\begin{table*}[t]
    \centering
    \caption{\textbf{Evaluation on Novel View Synthesis}. The \colorbox{first}{\textbf{best}}, \colorbox{second}{\underline{second best}} and \colorbox{third}{third best} results are highlighted for clarity. $\dagger$ indicates our reimplemented version of GS-LRM~\citep{zhang2024gs}. ``Ours (static)'' refers to our method pretrained solely on static datasets without the motion head. The same camera parameters are provided for all methods for fair comparison.}
    \label{tab:nvs}
    \vspace{-0.2cm}
    \renewcommand\arraystretch{1}
    \resizebox{\textwidth}{!}{

    \begin{tabular}{l|c|ccc|ccc|ccc}
        \toprule

        \multicolumn{1}{c|}{\multirow{2}{*}{\makecell[c]{Novel \\ View Synthesis}}} & \multicolumn{1}{c|}{\multirow{2}{*}{\makecell[c]{Time\\Per Scene}}} & \multicolumn{3}{c|}{RealEstate10K} & \multicolumn{3}{c|}{DyCheck} & \multicolumn{3}{c}{NVIDIA} \\

        \cmidrule(lr){3-5} \cmidrule(lr){6-8} \cmidrule(lr){9-11} &

        & $\uparrow$ PSNR &
        $\uparrow$ SSIM &
        $\downarrow$ LPIPS &
        $\uparrow$ mPSNR &
        $\uparrow$ mSSIM &
        $\downarrow$ mLPIPS &
        $\uparrow$ PSNR &
        $\uparrow$ SSIM &
        $\downarrow$ LPIPS \\

        \midrule

        \textit{Static Feed-forward} & & & & & & & & & & \\
        DepthSplat~\citep{xu2024depthsplat} & \color{lightgray}{0.60s} & 26.57 & \cellcolor{third}80.06 & \cellcolor{third}0.124 & 13.83 & 43.64 & 0.3850 & 17.16 & 50.02 & 0.3023 \\
        GS-LRM$^\dagger$~\citep{zhang2024gs} & \color{lightgray}{0.57s} & \cellcolor{third}26.94 & 79.13 & 0.139 & 14.60 & 45.35 & 0.3775 & 17.83 & 49.88 & 0.3265 \\
        Ours (static) & \color{lightgray}{0.84s} & \cellcolor{first}\textbf{27.60} & \cellcolor{second}\underline{81.25} & \cellcolor{second}\underline{0.113} & 15.24 & 47.84 & 0.3783 & \cellcolor{third}18.73 & \cellcolor{third}50.42 & \cellcolor{second}\underline{0.2959} \\

        \midrule

        \textit{Optimization-based} & & & & & & & & & & \\
        Splatter-a-Video~\citep{sun2024splatter} & \cellcolor{third}37min & - & - & - & 13.61 & 31.31 & 0.5706 & 14.39 & 25.38 & 0.5983 \\
        Shape-of-Motion~\citep{wang2024shape} & \cellcolor{second}\underline{10min} & - & - & - & \cellcolor{third}17.96 & \cellcolor{second}\underline{56.62} & \cellcolor{second}\underline{0.3463} & 15.30 & 31.69 & 0.5087 \\
        MoSca~\citep{lei2025mosca} & 45min & - & - & - & \cellcolor{second}\underline{18.24} & \cellcolor{third}55.14 & \cellcolor{third}0.3698 & \cellcolor{first}\textbf{21.45} & \cellcolor{first}\textbf{71.23} & \cellcolor{first}\textbf{0.2653} \\

        \midrule

        \cellcolor{mygray}Ours & \cellcolor{first}\textbf{0.93s} & \cellcolor{second}\underline{26.98} & \cellcolor{first}\textbf{81.75} & \cellcolor{first}\textbf{0.111} & \cellcolor{first}\textbf{18.46} & \cellcolor{first}\textbf{58.87} & \cellcolor{first}\textbf{0.3094} & \cellcolor{second}\underline{19.16} & \cellcolor{second}\underline{51.41} & \cellcolor{third}0.3152 \\

        \bottomrule
    \end{tabular}

    }
\end{table*}

\begin{table*}[t]
    \centering
    \caption{\textbf{Evaluation on 3D Point Tracking}. The \colorbox{first}{\textbf{best}}, \colorbox{second}{\underline{second best}} and \colorbox{third}{third best} results are highlighted for clarity. $\dagger$ denotes combining a depth estimation model~\citep{chen2025video}. All methods are given the same camera information for unprojecting 3D tracked points.}
    \label{tab:tracking}
    \vspace{-0.2cm}
    \renewcommand\arraystretch{1}
    \resizebox{\textwidth}{!}{

    \begin{tabular}{l|ccc|ccc|ccc}
        \toprule

        \multicolumn{1}{c|}{\multirow{2}{*}{\makecell[c]{3D Point \\ Tracking}}} & \multicolumn{3}{c|}{Aria Digital Twin} & \multicolumn{3}{c|}{DriveTrack} & \multicolumn{3}{c}{Panoptic Studio} \\

        \cmidrule(lr){2-4} \cmidrule(lr){5-7} \cmidrule(lr){8-10} &

        $\downarrow$ EPE$_{3D}$ &
        $\uparrow$ $\delta^{0.05}_{3D}$ &
        $\uparrow$ $\delta^{0.10}_{3D}$ &
        $\downarrow$ EPE$_{3D}$ &
        $\uparrow$ $\delta^{0.05}_{3D}$ &
        $\uparrow$ $\delta^{0.10}_{3D}$ &
        $\downarrow$ EPE$_{3D}$ &
        $\uparrow$ $\delta^{0.05}_{3D}$ &
        $\uparrow$ $\delta^{0.10}_{3D}$ \\

        \midrule

        BootsTAPIR~\citep{doersch2024bootstap}$^\dagger$ & \cellcolor{third}0.5539 & 17.73\% & 32.97\% & 0.0617 & \cellcolor{third}55.82\% & 75.66\% & 0.0650 & \cellcolor{third}69.28\% & 87.95\% \\
        CoTracker3~\citep{karaev2024cotracker}$^\dagger$ & 0.5614 & \cellcolor{second}\underline{19.88\%} & \cellcolor{third}35.82\% & \cellcolor{third}0.0637 & 55.30\% & \cellcolor{third}77.55\% & \cellcolor{third}0.0617 & 69.27\% & \cellcolor{third}88.04\% \\
        SpatialTracker~\citep{xiao2024spatialtracker} & \cellcolor{second}\underline{0.5413} & \cellcolor{third}18.08\% & \cellcolor{second}\underline{38.23\%} & \cellcolor{second}0.0648 & \cellcolor{second}\underline{56.58\%} & \cellcolor{first}\textbf{80.67\%} & \cellcolor{second}\underline{0.0519} & \cellcolor{second}\underline{72.91\%} & \cellcolor{second}89.86\% \\

        \cellcolor{mygray}Ours & \cellcolor{first}\textbf{0.2153} & \cellcolor{first}\textbf{52.05\%} & \cellcolor{first}\textbf{71.63\%} & \cellcolor{first}\textbf{0.0472}& \cellcolor{first}\textbf{60.63\%} & \cellcolor{second}\underline{79.87\%} & \cellcolor{first}\textbf{0.0352} & \cellcolor{first}\textbf{87.88\%} & \cellcolor{first}\textbf{94.61\%} \\

        \bottomrule
    \end{tabular}

    }
\end{table*}

\begin{figure*}[t]
    \begin{center}
    \includegraphics[width=0.89\textwidth]{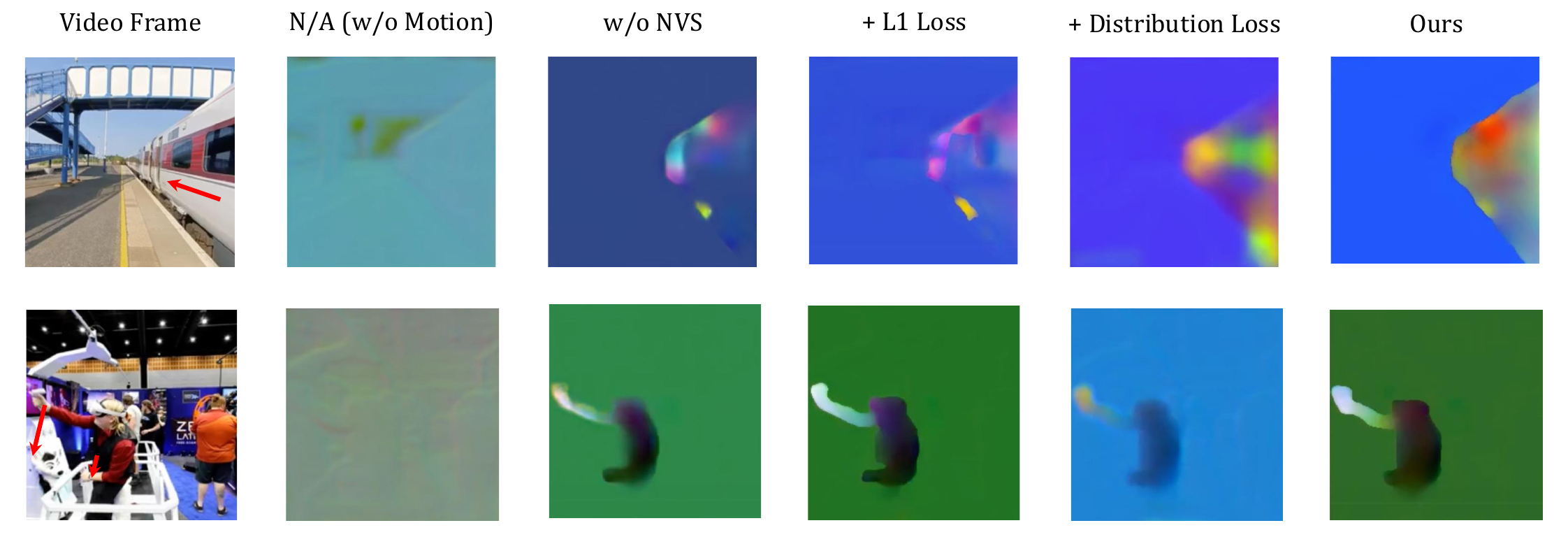}
    \end{center}
    \vspace{-0.4cm}
    \caption{\textbf{Motion Visualization for Ablation Studies}. We investigate key factors affecting motion learning in \name, such as loss design and synergy with view synthesis. XYZ values in motion maps are normalized as RGB for visualization. Red arrows on video frames indicate motion directions.}
    \label{fig:motion_vis}
\end{figure*}

\begin{table*}[t]
    \vspace{-0.2cm}
    \begin{minipage}[t]{0.475\textwidth}
        \caption{\textbf{Ablation study} on camera conditioning.}
        \vspace{-0.2cm}
        \label{tab:ablation_camera}
        \renewcommand\arraystretch{1}
        \resizebox{\textwidth}{!}{
            \begin{tabular}{l|cccc}
                \toprule

                \multicolumn{1}{c|}{\multirow{2}{*}{\makecell[c]{Camera\\Conditioning}}} & \multicolumn{3}{c}{RealEstate10K} \\

                \cmidrule(lr){2-4} &

                $\uparrow$ PSNR &
                $\uparrow$ SSIM &
                $\downarrow$ LPIPS \\

                \midrule

                N/A & 25.56 & 74.13 & 0.150 \\
                + Pl\"ucker embedding & 25.81 & 74.44 & 0.143 \\
                + Camera token & \underline{26.81} & \underline{78.65} & \underline{0.121} \\
                \cellcolor{mygray}Ours & \cellcolor{mygray}\textbf{27.60} & \cellcolor{mygray}\textbf{81.25} & \cellcolor{mygray}\textbf{0.113} \\

                \bottomrule
            \end{tabular}
        }
    \end{minipage}
    \hfill
    \begin{minipage}[t]{0.475\textwidth}
        \caption{\textbf{Ablation study} on motion supervision.}
        \vspace{-0.2cm}
        \label{tab:ablation_motion}
        \renewcommand\arraystretch{0.98}
        \resizebox{\textwidth}{!}{
            \begin{tabular}{l|cccc}
                \toprule

                \multicolumn{1}{c|}{\multirow{2}{*}{\makecell[c]{Motion\\Supervision}}} & \multicolumn{3}{c}{Aria Digital Twin} \\

                \cmidrule(lr){2-4} &

                $\downarrow$ EPE$_{3D}$ &
                $\uparrow$ $\delta^{0.05}_{3D}$ &
                $\uparrow$ $\delta^{0.10}_{3D}$ \\

                \midrule

                N/A & 0.7938 & 19.58\% & 32.86\% \\
                + Point-wise L1 & \underline{0.2262} & \underline{48.74\%} & \underline{69.93\%} \\
                + Distribution loss & 0.2496 & 45.98\% & 66.87\% \\
                \cellcolor{mygray}Ours & \cellcolor{mygray}\textbf{0.2153} & \cellcolor{mygray}\textbf{52.05\%} & \cellcolor{mygray}\textbf{71.63\%} \\

                \bottomrule
            \end{tabular}
        }
    \end{minipage}
    \vspace{-0.3cm}
\end{table*}

\begin{table*}[t]
    \centering
    \caption{\textbf{Ablation Study} on the synergy of motion estimation and novel view synthesis (NVS).}
    \vspace{-0.2cm}
    \label{tab:ablation_synergy}
    \renewcommand\arraystretch{1}
    \resizebox{\textwidth}{!}{

    \begin{tabular}{l|ccc|ccc|ccc}
        \toprule

        \multicolumn{1}{c|}{\multirow{2}{*}{\makecell[c]{Motion \&\\View Synthesis}}} & \multicolumn{3}{c|}{DyCheck} & \multicolumn{3}{c|}{NVIDIA} & \multicolumn{3}{c}{Aria Digital Twin} \\

        \cmidrule(lr){2-4} \cmidrule(lr){5-7} \cmidrule(lr){8-10} &

        $\uparrow$ mPSNR &
        $\uparrow$ mSSIM &
        $\downarrow$ mLPIPS &
        $\uparrow$ PSNR &
        $\uparrow$ SSIM &
        $\downarrow$ LPIPS &
        $\downarrow$ EPE$_{3D}$ &
        $\uparrow$ $\delta^{0.05}_{3D}$ &
        $\uparrow$ $\delta^{0.10}_{3D}$ \\

        \midrule

        NVS w/o motion & 15.82 & \underline{47.96} & \underline{0.3741} & 18.38 & 48.56 & \textbf{0.3010} & 0.7938 & 19.58\% & 32.86\% \\
        Motion w/o NVS & \underline{16.26} & 45.56 & 0.3461 & \underline{18.98} & \underline{49.39} & 0.3207 & \underline{0.3801} & \underline{24.72\%} & \underline{42.92\%} \\
        \cellcolor{mygray}Ours & \cellcolor{mygray}\textbf{18.46} & \cellcolor{mygray}\textbf{58.87} & \cellcolor{mygray}\textbf{0.3094} & \cellcolor{mygray}\textbf{19.16} & \cellcolor{mygray}\textbf{50.41} & \cellcolor{mygray}\underline{0.3152} & \cellcolor{mygray}\textbf{0.2153} & \cellcolor{mygray}\textbf{52.05\%} & \cellcolor{mygray}\textbf{71.63\%} \\

        \bottomrule
    \end{tabular}

    }
\end{table*}

\begin{figure*}[t]
    \vspace{-0.4cm}
    \begin{center}
    \includegraphics[width=0.9\textwidth]{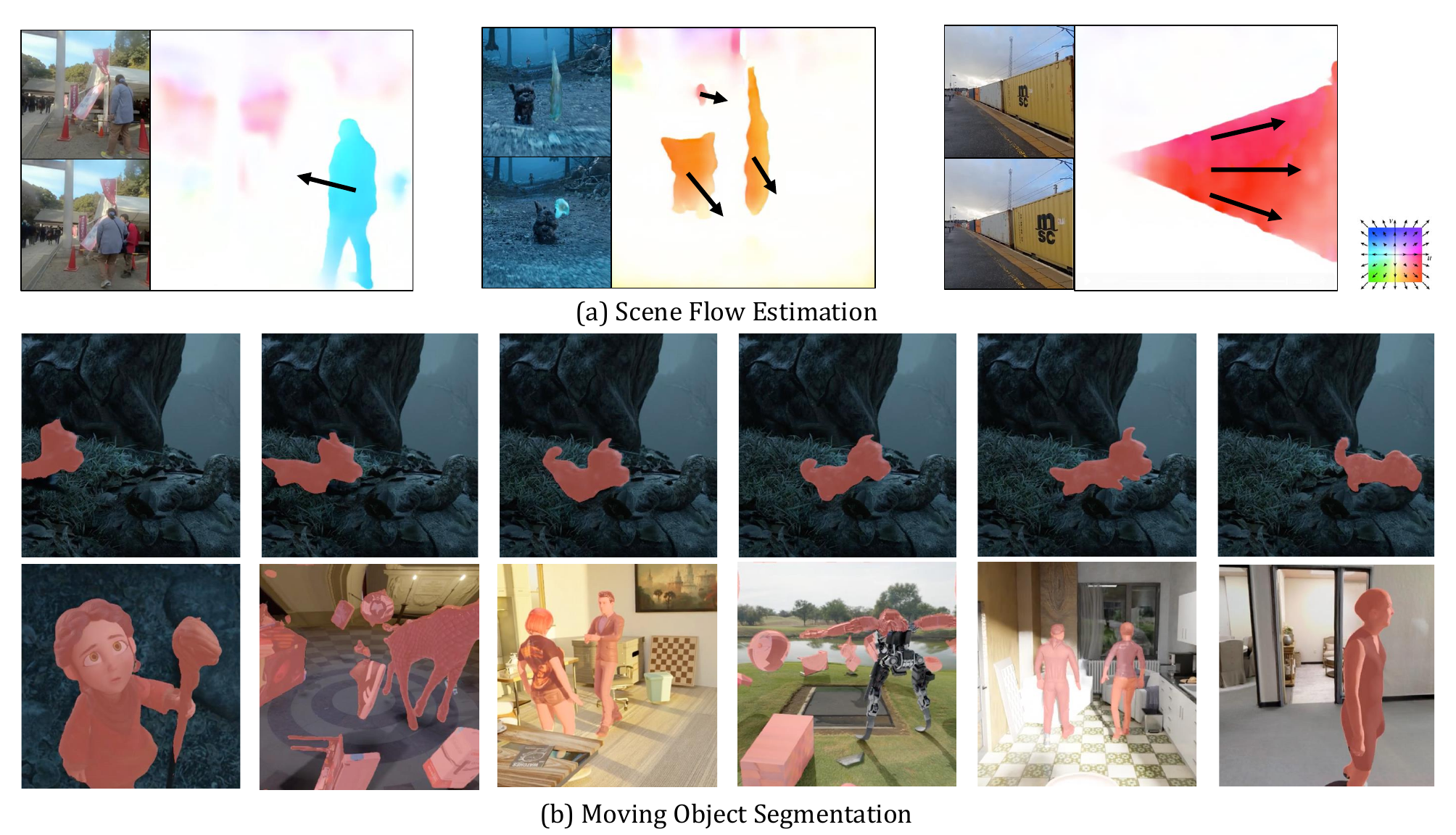}
    \end{center}
    \vspace{-0.6cm}
    \caption{\textbf{Zero-shot Applications}. Predicted motion maps from our model can be directly applied to downstream tasks, such as (a) scene flow estimation and (b) moving object segmentation, in a \textbf{zero-shot} manner, without any task-specific fine-tuning or supervision.}
    \label{fig:app}
    \vspace{-0.4cm}
\end{figure*}

\subsection{Experimental Settings}\label{subsec:settings}

\paragraph{Implementation}
\name is built on a geometrically pretrained transformer, VGGT~\citep{wang2025vggt}, via full fine-tuning.
The splatter head and camera/time embeddings are trained from scratch.
These new embeddings are zero-initialized to provide a stable initialization without interfering with the pretrained model.
AdamW~\citep{loshchilov2017decoupled} with cosine learning rate scheduling and linear warm-up is used for optimization.
We observed that training \name is particularly unstable, likely due to sparse annotations and the heterogeneous nature of training datasets.
A curriculum strategy is used that gradually increases the training complexity, involving (1) pretraining on static scenes, (2) dynamic scenes with varying views, and (3) fine-tuning in high resolution.
Several techniques, such as \texttt{gsplat} rendering backend~\citep{ye2025gsplat}, DeepSpeed~\citep{rasley2020deepspeed}, gradient checkpointing~\citep{chen2016training}, gradient accumulation, and \texttt{bf16} mixed precision, are adopted to improve memory and computation efficiency.
Training completes in approximately 5 days using 32 H20 GPUs.
More details are provided in the supplementary material.

\vspace{-0.4cm}
\paragraph{Evaluation}
We evaluate \name on two primary tasks: novel view synthesis (Sec.~\ref{subsec:4dvs}) and 3D point tracking (Sec.~\ref{subsec:3dtrack}).
Considering pose-aware scenarios, we provide the same camera parameters for all methods to ensure a fair comparison.
Following prior works~\citep{charatan2024pixelsplat,chen2024mvsplat,xu2024depthsplat,wang2024shape,sun2024splatter,lei2025mosca,liang2024feed}, RealEstate10K~\citep{zhou2018stereo} is used to evaluate novel view synthesis performance on static scenes.
DyCheck~\citep{gao2022dynamic} and NVIDIA dynamic scene dataset~\citep{yoon2020novel} are adopted for dynamic scenes, in terms of PSNR, SSIM and LPIPS~\citep{zhang2018perceptual}.
Since the DyCheck dataset contains invisible regions in the target novel views, we compute reconstruction metrics using the provided covisibility masks (denoted by a prefix ``m''), ensuring a fair comparison across methods.
For 3D point tracking, the TAPVid-3D~\citep{koppula2024tapvid} benchmark serves as the evaluation protocol, covering both indoor and outdoor scenes across three datasets.
End-point Euclidean error in the 3D space (EPE$_{3D}$) and the percentage of 3D points within $0.05$ and $0.10$ units of the ground-truth locations ($\delta^{0.05}_{3D}$ and $\delta^{0.10}_{3D}$) are utilized as metrics.
A large pretrained video depth estimation model, Video Depth Anything~\citep{chen2025video}, and the same camera information are combined to unproject the predicted 2D tracking points of baselines into the 3D space for fair comparison.

\subsection{Novel View Synthesis}\label{subsec:4dvs}

\subsubsection{Static Scene View Synthesis}
As a special case of dynamic scenes, we first evaluate \name on a static dataset, RealEstate10K \citep{zhou2018stereo}, comparing against several state-of-the-art feed-forward static scene reconstruction methods, including DepthSplat~\citep{xu2024depthsplat}, our reimplementation of GS-LRM~\citep{zhang2024gs}, and our method pretrained solely on the first-stage static reconstruction.
As shown in Table~\ref{tab:nvs}, although \name is primarily designed for dynamic scenes, it maintains competitive performance on static scenes.
Notably, when processing static inputs, our predicted motion naturally converges to \textit{zero} (less than 1e-3), demonstrating the ability of \name to implicitly differentiate between static and dynamic regions.

\subsubsection{Dynamic Scene View Synthesis}
We compare against three state-of-the-art open-sourced methods~\citep{wang2024shape,sun2024splatter,lei2025mosca} for dynamic 3D Gaussian reconstruction, and also report static feed-forward baselines for reference.
For DyCheck, we sample 13 frames of size $518 \times 518$ from a random clip of 65 frames from iPhone-record videos, and include the other two cameras for evaluation.
For NVIDIA, 12 input views at $379 \times 672$ resolution are sampled in a round-robin style~\citep{gao2021dynamic,liu2023robust,liang2024feed}, where the \texttt{i}-th frame comes from the \texttt{i}-th camera at timestep \texttt{i}.

As shown in Table~\ref{tab:nvs}, \name achieves competitive or superior performance compared to baselines, while requiring only \texttt{0.93}s per scene, which is orders of magnitude faster than prior approaches that rely on heavy pretrained models and complex multi-stage pipelines.
Qualitative visualizations in Figure~\ref{fig:nvs_vis} further highlight the strength of our approach.
While MoSca~\citep{lei2025mosca} demonstrates impressive performance, it struggles with sparse inputs and often overfits to seen poses, producing spiky and over-saturated artifacts under novel views and timesteps.
In contrast, \name leverages large-scale learned priors to generalize more effectively, yielding smoother and more realistic results.

To ensure fair comparison and better reflect real scenarios, no video masks for dynamic objects are used in our experiments.
It poses a major challenge for optimization-based methods like Shape-of-Motion~\citep{wang2024shape} and Splatter-a-Video~\citep{sun2024splatter}, which rely heavily on explicit motion segmentation.
The challenge is particularly evident on the NVIDIA dataset, where significant camera shake hinders disentangling scene dynamics, leading to notably degraded performance and even worse than static baselines.
In contrast, \name exhibits strong robustness by directly learning to model motion.
To evaluate the robustness of our method, visualization results on the \textit{in-the-wild} DAVIS dataset~\citep{perazzi2016davis} are provided in the supplementary material, where camera poses are estimated by MegaSaM~\citep{li2025megasam}.

\vspace{-0.1cm}
\subsection{3D Point Tracking}\label{subsec:3dtrack}
Trained on large-scale point tracking datasets~\citep{zheng2023pointodyssey,karaev2023dynamicstereo,jin2024stereo4d}, the proposed method can also \textit{densely} track any 3D point corresponding to a pixel across video frames.
We compare \name against three strong baselines, including two state-of-the-art 2D point tracking methods, BootsTAP~\citep{doersch2024bootstap} and CoTracker3~\citep{karaev2024cotracker}, as well as a native 3D point tracking approach, SpatialTracker~\citep{xiao2024spatialtracker}.
For 2D trackers, a recent video depth estimation model~\citep{chen2025video} and ground-truth camera intrinsics are used to unproject tracked points into 3D space.
Considering scale differences across methods, we normalize all predicted 3D points by their median norm before evaluation.

Quantitative results are reported in Table~\ref{tab:tracking}.
While 3D-based SpatialTracker generally outperforms 2D-based approaches, all of them rely heavily on pretrained monocular depth estimators for geometry reasoning, introducing significant noise and inconsistency in the 3D space.
In contrast, \name directly estimates 3D point positions in a \textit{shared} world coordinate, enabling more accurate and robust 3D tracking, and achieves consistently superior or competitive performance across all datasets.
Visualization results are provided in the supplementary material.

\subsection{Ablation and Analysis}\label{subsec:ablation}

\paragraph{Camera Conditioning}
We investigate different camera conditioning strategies in the static pretraining stage.
As shown in Table~\ref{tab:ablation_camera}, camera tokens are injected throughout the feature backbone, enabling effective camera-aware modeling.
In contrast, Pl\"ucker embeddings provide limited conditioning on their own and are merely comparable to having no camera information at all.
However, as a pixel-aligned representation, Pl\"ucker embeddings are complementary to camera tokens, and their combination yields the most effective camera conditioning.

\vspace{-0.4cm}
\paragraph{Motion Supervision}
To learn 3D movements in dynamic scenes, we provide two kinds of motion supervision: (1) point-wise L1 loss and (2) distribution loss, as described in Equation~\ref{equ:motion}.
Their efficiency is evaluated via the 3D point tracking task in Table~\ref{tab:ablation_motion}.
Without any motion supervision, i.e., learning solely from novel view synthesis, training exhibits severe loss oscillations and frequent \texttt{None} gradients. The distribution loss captures only relative motion between pixels, while the point-wise L1 loss produces more reasonable motion maps. Combining both leads to sharper boundaries.
Qualitative results of estimated motions from different motion objectives are provided in Figure~\ref{fig:motion_vis}.

\vspace{-0.4cm}
\paragraph{Synergy of Motion and View Synthesis}
Thanks to the unified design of \name, it supports simultaneous novel view synthesis (NVS) and motion estimation.
We study their mutual benefits in Table~\ref{tab:ablation_synergy}.
``NVS w/o motion'' disables explicit motion supervision during training, relying solely on NVS as a proxy to learn dynamics.
As shown in Table~\ref{tab:ablation_synergy} and Figure~\ref{fig:motion_vis}, this setting fails to learn meaningful motion and tends to model static scenes.
``Motion w/o NVS'' detaches the motion head from 3DGS rendering and instead conditions the depth head on time.
Although explicit supervision enables some motion learning, the predictions are blurry and low-quality, as shown in Figure~\ref{fig:motion_vis}.
Moreover, the depth head must now model both geometry and dynamics, increasing its burden and negatively affecting NVS.
These results highlight the mutual reinforcement between NVS and motion estimation, where joint training leads to better performance on both.

\vspace{-0.4cm}
\paragraph{VGGT Initialization}
Although \name is initialized by a pretrained model, i.e., VGGT~\citep{wang2025vggt}, it is not an essential component of the proposed method.
We verified it in our preliminary experiments that using a feature backbone trained entirely from scratch still leads to comparable performance with about 3 times slower convergence.
The pretrained VGGT backbone simply serves to accelerate training, but does not fundamentally change the outcome.
Specifically, we conduct quantitative experiments on DyCheck at $224\times 224$ resolution.
\name initialized from VGGT reaches 17.97 dB mPSNR within 2 days, whereas the counterpart trained from scratch achieves 16.64 dB after 6 days and has not yet fully converged, suggesting that continuous training could further improve its performance.

\vspace{-0.1cm}
\subsection{Zero-shot Applications}\label{subsec:app}

\paragraph{Scene Flow Estimation}
Scene flow can be naturally derived by transforming the estimated per-pixel motion vectors from world coordinates to the target camera's coordinates.
Visualization results in Figure~\ref{fig:app}(a) present sharp edges and accurate motion directions.
Different colors in the flow maps indicate different movement directions that are marked by block arrows.
More visualization results are provided in the supplementary material.

\vspace{-0.4cm}
\paragraph{Moving Object Segmentation}
By thresholding the norm of per-pixel motion vectors, the estimated motion maps can also be used to segment moving objects (Figure~\ref{fig:app}(b)), which is an essential task in computer vision and robotics~\citep{xie2024moving,huang2025segment}.
Remarkably, this is achieved without any explicit supervision during training, demonstrating the strong potential of our method.
More visualization results are provided in the supplementary material.

\section{Conclusion}\label{sec:conclu}
We introduce \name, a feed-forward model for dynamic novel view synthesis from monocular videos.
It's trained on large-scale, diverse datasets and jointly models scene appearance, geometry, and motion in a unified and efficient framework.
Dynamic splatter pixels are proposed to represent dynamic scenes, enabling accurate and temporally coherent 4D reconstruction.
\name also supports various applications, such as depth estimation, 3D point tracking, scene flow estimation, and moving object segmentation, showcasing its versatility for dynamic scene perception.
We hope this work could serve as a step toward generalizable dynamic scene understanding, and broader applications requiring spatial intelligence.
Limitations and future work are discussed in the supplementary material.

{
    \small
    \bibliographystyle{ieeenat_fullname}
    \bibliography{references}
}

\appendix
\setcounter{figure}{0}
\setcounter{table}{0}
\renewcommand{\thefigure}{S.\arabic{figure}}
\renewcommand{\thetable}{S.\arabic{table}}
\renewcommand{\theequation}{S.\arabic{equation}}
\maketitlesupplementary

\section{Implementation Details}\label{apx:impl}

\subsection{Dataset Details}
\begin{table*}[t]
    \centering
    \caption{\textbf{Training Datasets}. Eight datasets from diverse sources are utilized to train \name at scale.
    ``\#Repeat'' denotes dataset duplication count during integration to balance their contributions.}
    \label{tab:datasets_full}
    \renewcommand\arraystretch{1}

    \begin{tabular}{lccccrrr}
        \toprule

        \textbf{Dataset} & \textbf{Dynamic?} & \textbf{Depth?} & \textbf{Tracking?} & \textbf{Real?} & \textbf{\#Scenes} & \textbf{\#Frames} & \textbf{\#Repeat} \\

        \midrule

        RealEstate10K~\citep{zhou2018stereo} & \no & \no & \no & \ye & 70K & 6.36M & 1$\times$ \\
        TartanAir~\citep{wang2020tartanair} & \no & \ye & \no & \no & 0.4K & 0.49M & 100$\times$ \\
        MatrixCity~\citep{li2023matrixcity} & \no & \ye & \no & \no & 4.5K & 0.31M & 10$\times$ \\
        PointOdyssey~\citep{zheng2023pointodyssey} & \ye & \ye & \ye & \no & 0.1K & 0.18M & 1000$\times$ \\
        DynamicReplica~\cite{karaev2023dynamicstereo} & \ye & \ye & \ye & \no & 0.5K & 0.26M & 100$\times$ \\
        Spring~\citep{mehl2023spring} & \ye & \ye & \no & \no & 0.03K & 0.003M & 2000$\times$ \\
        VKITTI2~\citep{cabon2020virtual} & \ye & \ye & \no & \no & 0.1K & 0.03M & 500$\times$ \\
        Stereo4D~\citep{jin2024stereo4d} & \ye & \ye & \ye & \ye & 98K & 19.6M & 1$\times$ \\

        \bottomrule
    \end{tabular}

\end{table*}

Detailed information about the training datasets used in this work is provided in Table~\ref{tab:datasets_full}.
The majority of training datasets used in our experiments are synthetic, providing rich annotations such as pixel-aligned depth maps and accurate camera intrinsics and extrinsics.
Ground-truth 3D point trajectories are also available in PointOdyssey~\citep{zheng2023pointodyssey} and DynamicReplica~\citep{karaev2023dynamicstereo}.
On the other hand, Stereo4D~\citep{jin2024stereo4d} is a large-scale dataset constructed from YouTube stereo videos and annotated using pretrained foundation models~\citep{teed2020raft,doersch2024bootstap}.
Despite being partially noisy, its diversity and scale offer strong generalization benefits.

The pretrained backbone, VGGT~\citep{wang2025vggt}, requires all 3D scenes to be normalized to a unit scale on average, which in turn necessitates depth maps and camera extrinsics to be defined in a consistent metric space.
It is satisfied by all synthetic datasets and Stereo4D, whose camera poses and depths are metric-aligned by construction.
However, RealEstate10K~\citep{zhou2018stereo} only provides relative camera parameters estimated via COLMAP~\citep{schonberger2016structure}, resulting in an unknown global scale.
To address this issue, we re-estimate both depth maps and camera extrinsics using recent foundation models, including Video Depth Anything~\citep{chen2025video}, Depth Pro~\citep{bochkovskii2024depth} and MegaSaM~\citep{li2025megasam}, to recover aligned geometry across frames.

\subsection{Curriculum Training}
We observed that training \name is particularly unstable: the training loss often fluctuates abruptly, and gradients are prone to becoming \texttt{None}.
It may arise from sparse annotations and the heterogeneous nature of training datasets, which mix datasets from various sources with differing domains (e.g., indoor \textit{vs}. outdoor, real \textit{vs}. synthetic), camera FoV, recording frame rates, etc.

Thanks to the versatile design of \name, we employ a curriculum strategy that gradually increases the training complexity.
It begins by pretraining the model on low-resolution ($224\times224$) static datasets with only depth and photometric losses, then introduces dynamic datasets along with motion supervision.
We found that static datasets play a crucial role in stabilizing training for dynamic scenes, without which the loss would be highly unstable.
Since modeling dynamic scenes requires reconstructing a set of 3DGS for each query time, which results in high GPU memory usage, we start by training on 5 input views for dynamic scenes and then expand to 13 views for fine-tuning.
Finally, the training resolution is increased to $518$.
Similar to VGGT~\citep{wang2025vggt}, in the last training stage, we randomly sample the frame number from $2\sim13$ and the aspect ratio from $0.5\sim2$, with the largest side fixed to $518$.

\subsection{Training Details}
Image encoder, feature backbone, and depth head of \name are initialized from VGGT~\citep{wang2025vggt}.
Motion head is initialized from its pointmap head.
Remaining components, such as the splatter head and camera/time embeddings, are trained from scratch.
We use AdamW optimizer~\citep{loshchilov2017decoupled} with a weight decay of \texttt{0.05}, and adopt a cosine learning rate scheduler~\citep{loshchilov2016sgdr} with linear warm-up for all curriculum training stages.
For static pretraining and dynamic scenes with 5 and 13 input views at a resolution of $224 \times 224$, we use learning rates of \texttt{4e-4}, \texttt{4e-4} and \texttt{4e-5}, respectively, with a batch size of \texttt{256}.
Training rates for parameters initialized from VGGT are multiplied by \texttt{0.1}.
With \texttt{32} H20 GPUs, training takes about \texttt{2} days for static and then dynamic scenes with 5 views, and \texttt{2} days for 13 views.
\name is then finetuned on $518\times518$ videos with 13 frames using a learning rate of \texttt{1e-5}, which takes around \texttt{1} day.
To improve memory and computation efficiency, several techniques such as \texttt{gsplat}~\citep{ye2025gsplat} rendering backend, DeepSpeed~\citep{rasley2020deepspeed}, gradient checkpointing~\citep{chen2016training}, gradient accumulation, and \texttt{bf16} mixed precision are also adopted.

For the multi-task training objective, we did not manually tune the relative weights between different loss terms.
Instead, the weights were set to bring the numerical values of the losses into roughly similar ranges: $\lambda_\text{d}=\lambda_\text{r}=\lambda_{\text{pt}}=1$ and $\lambda_\text{m}=\lambda_{\text{dist}}=10$.

\section{Limitations and Future Work}\label{apx:discussion}
Although \name achieves competitive performance with inference speeds that are orders of magnitude faster than optimization-based methods, there remains a noticeable gap in reconstruction quality.
This gap arises partly because many optimization-based approaches benefit from multiple pretrained models that preprocess input videos to provide richer and more accurate cues.
Incorporating such richer prior knowledge directly into \name represents a promising direction for future work to further improve reconstruction fidelity and robustness.

Currently, \name depends on off-the-shelf tools for camera parameter estimation, which adds an external dependency and may limit end-to-end optimization.
Seamlessly integrating camera pose estimation within the \name pipeline could simplify the overall workflow, reduce error accumulation, and enhance adaptability to diverse scenarios.

Moreover, scaling \name to handle long videos and achieve high-resolution rendering remains a challenge.
The computational cost and memory demand grow significantly with scene complexity and resolution, which constrains practical deployment.
Developing more compact and efficient dynamic scene representations, possibly through novel model architectures or sparse encoding strategies, is essential to push 4D reconstruction towards real-world applications.

Addressing these challenges will not only bridge the performance gap but also unlock the full potential of fast and high-quality dynamic scene reconstruction.

\section{License Information}\label{apx:license}
We employ several open-source implementations in our experimental comparisons, including: (1) DepthSplat~\citep{xu2024depthsplat}\footnote{\url{https://github.com/cvg/depthsplat/tree/main}} (MIT License), (2) Splatter-a-Video~\citep{sun2024splatter}\footnote{\url{https://github.com/SunYangtian/Splatter_A_Video}} (Apache License), (3) Shape-of-Motion~\citep{wang2024shape}\footnote{\url{https://github.com/vye16/shape-of-motion/}} (MIT License), (4) MoSca~\citep{lei2025mosca}\footnote{\url{https://github.com/JiahuiLei/MoSca}} (MIT License), (5) BootsTAPIR~\citep{doersch2024bootstap}\footnote{\url{https://github.com/google-deepmind/tapnet}} (Apache License), (6) CoTracker3~\citep{karaev2024cotracker}\footnote{\url{https://github.com/facebookresearch/co-tracker}} (Creative Commons Attribution-NonCommercial 4.0 International Public License), and (7) SpatialTracker~\citep{li2024spacetime}\footnote{\url{https://github.com/henry123-boy/SpaTracker}} (Attribution-NonCommercial 4.0 International).

Datasets from diverse sources are utilized to train \name, including: (1) RealEstate10K~\citep{zhou2018stereo}\footnote{\url{https://google.github.io/realestate10k/}} (Creative Commons Attribution 4.0 International License), (2) TartanAir~\citep{wang2020tartanair}\footnote{\url{https://theairlab.org/tartanair-dataset/}} (Creative Commons Attribution 4.0 International License), (3) MatrixCity~\citep{li2023matrixcity}\footnote{\url{https://city-super.github.io/matrixcity/}} (Apache License), (4) PointOdyssey~\citep{zheng2023pointodyssey}\footnote{\url{https://pointodyssey.com/}} (MIT License), (5) DynamicReplica~\cite{karaev2023dynamicstereo}\footnote{\url{https://github.com/facebookresearch/dynamic_stereo}} (Attribution-NonCommercial 4.0 International License), (6) Spring~\citep{mehl2023spring}\footnote{\url{https://spring-benchmark.org/}} (CC BY 4.0), (7) VKITTI2~\citep{cabon2020virtual}\footnote{\url{https://europe.naverlabs.com/research/computer-vision/proxy-virtual-worlds-vkitti-2/}} (Creative Commons Attribution-NonCommercial-ShareAlike 3.0), and (8) Stereo4D~\citep{jin2024stereo4d}\footnote{\url{https://stereo4d.github.io/}} (CC0 1.0 Universal).

\section{Broader Impact}
\name provides substantial advantages for fields including robotics simulation, AR/VR, autonomous driving, and digital twins by enabling fast and generalizable dynamic scene understanding.
Its ability to efficiently reconstruct dynamic environments can accelerate innovation and improve system performance in these areas.
However, such powerful technology also raises risks related to the unauthorized generation of content and potential privacy violations.
To mitigate these risks, it is essential to establish clear ethical guidelines and implement appropriate regulatory measures to ensure responsible and safe usage.

\section{More Visualization Results}

\begin{figure*}[t]
    \begin{center}
    \includegraphics[width=0.84\textwidth]{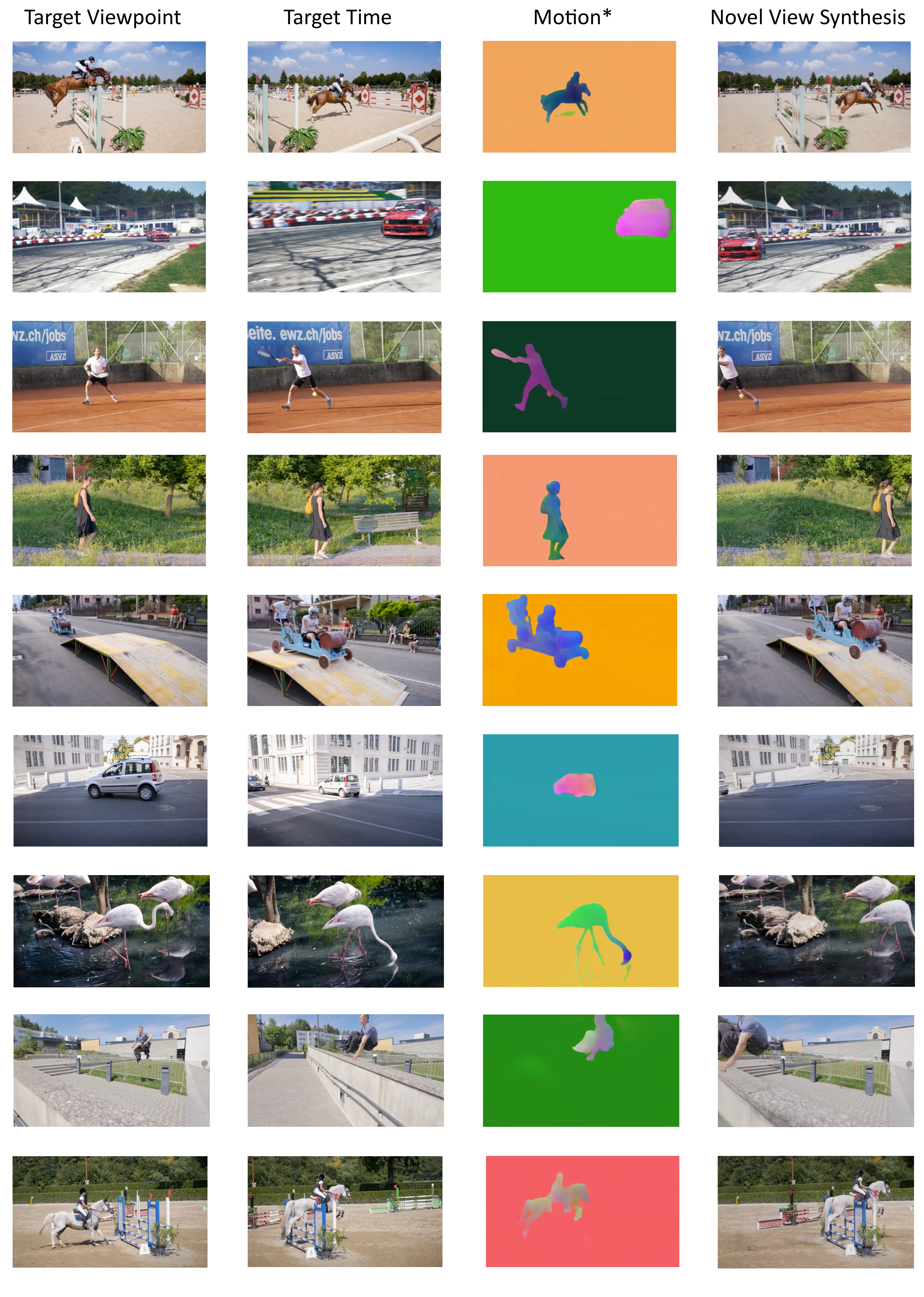}
    \end{center}
    \caption{\textbf{Qualitative results of novel view synthesis on the DAVIS dataset}~\citep{perazzi2016davis}. Camera poses are estimated by MegaSaM~\citep{li2025megasam}. ``Motion$^*$'' denotes the 3D pixel displacements between the ``Target Time'' frame with respect to the ``Target Viewpoint'' frame. The ``Novel View Synthesis'' results show the reconstructed scene at the target time from the target viewpoint.}
    \label{fig:nvs}
\end{figure*}

\begin{figure*}[t]
    \begin{center}
    \includegraphics[width=0.72\textwidth]{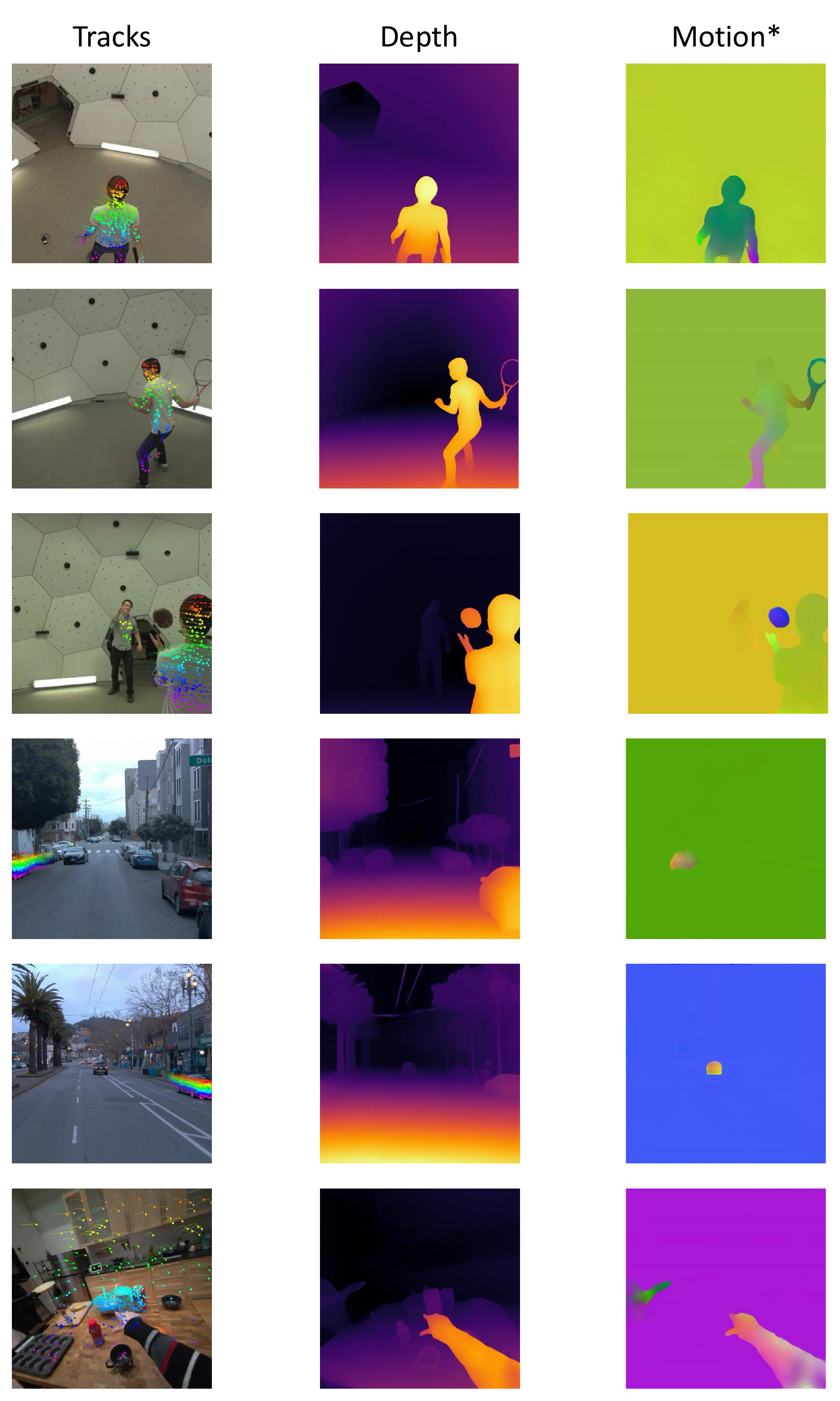}
    \end{center}
    \caption{\textbf{Qualitative results for 3D point tracking}. ``Motion$^*$'' means the 3D points' movements of the input frame with respect to the first one. Points in the 3D space are projected to the image space for 2D point tracking visualization.}
    \label{fig:track}
\end{figure*}

\begin{figure*}[t]
    \begin{center}
    \includegraphics[width=0.72\textwidth]{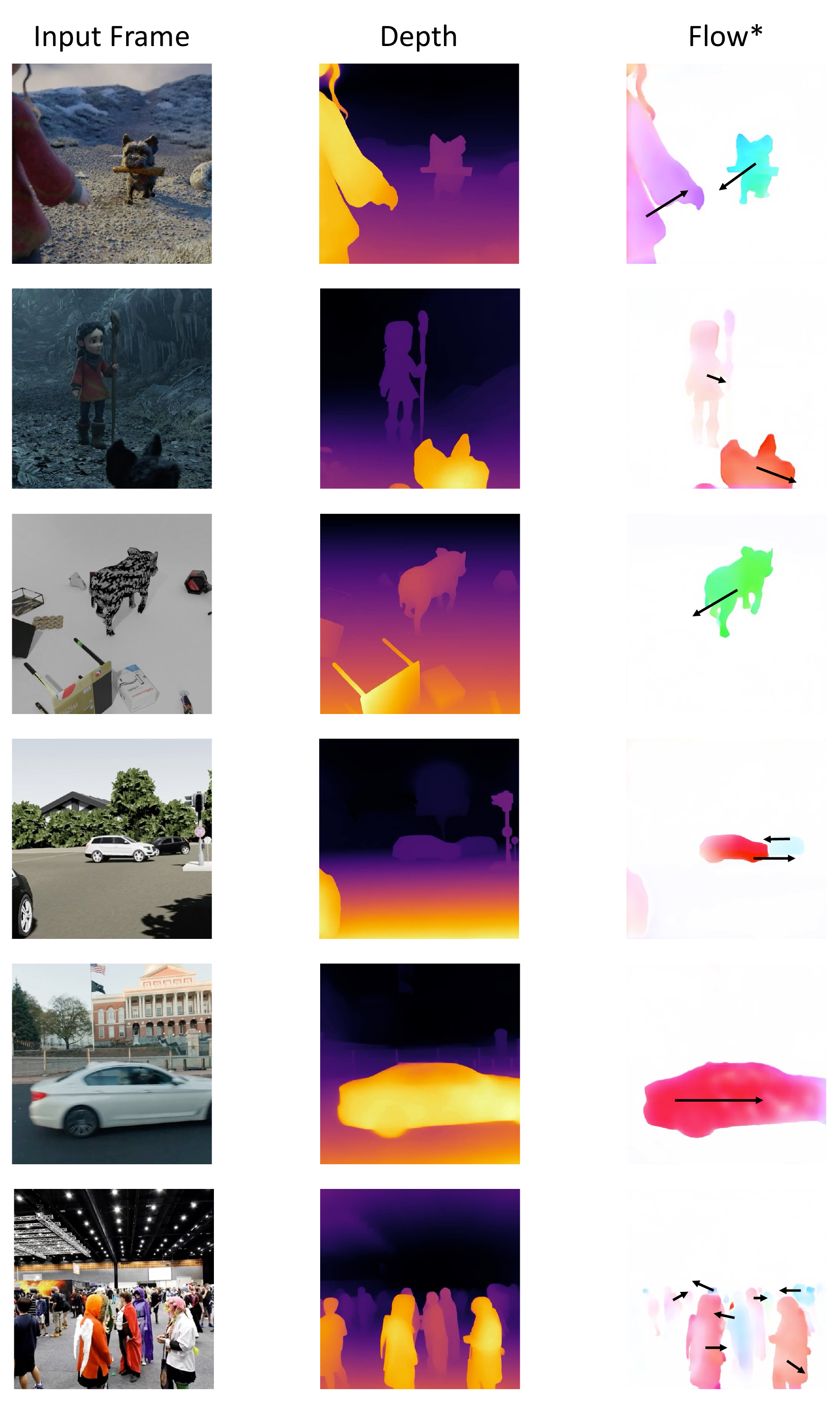}
    \end{center}
    \vspace{-0.3cm}
    \caption{\textbf{Qualitative results for scene flow estimation}. ``Flow$^*$'' means the optical flow of the input pixels with respect to the first frame, and is obtained by projecting the estimated motion maps in the 3D space to the camera space. Each color of optical flow means a specific 2D direction and arrows on pictures roughly mark the directions.}
    \label{fig:flow}
\end{figure*}

\begin{figure*}[t]
    \begin{center}
    \includegraphics[width=0.72\textwidth]{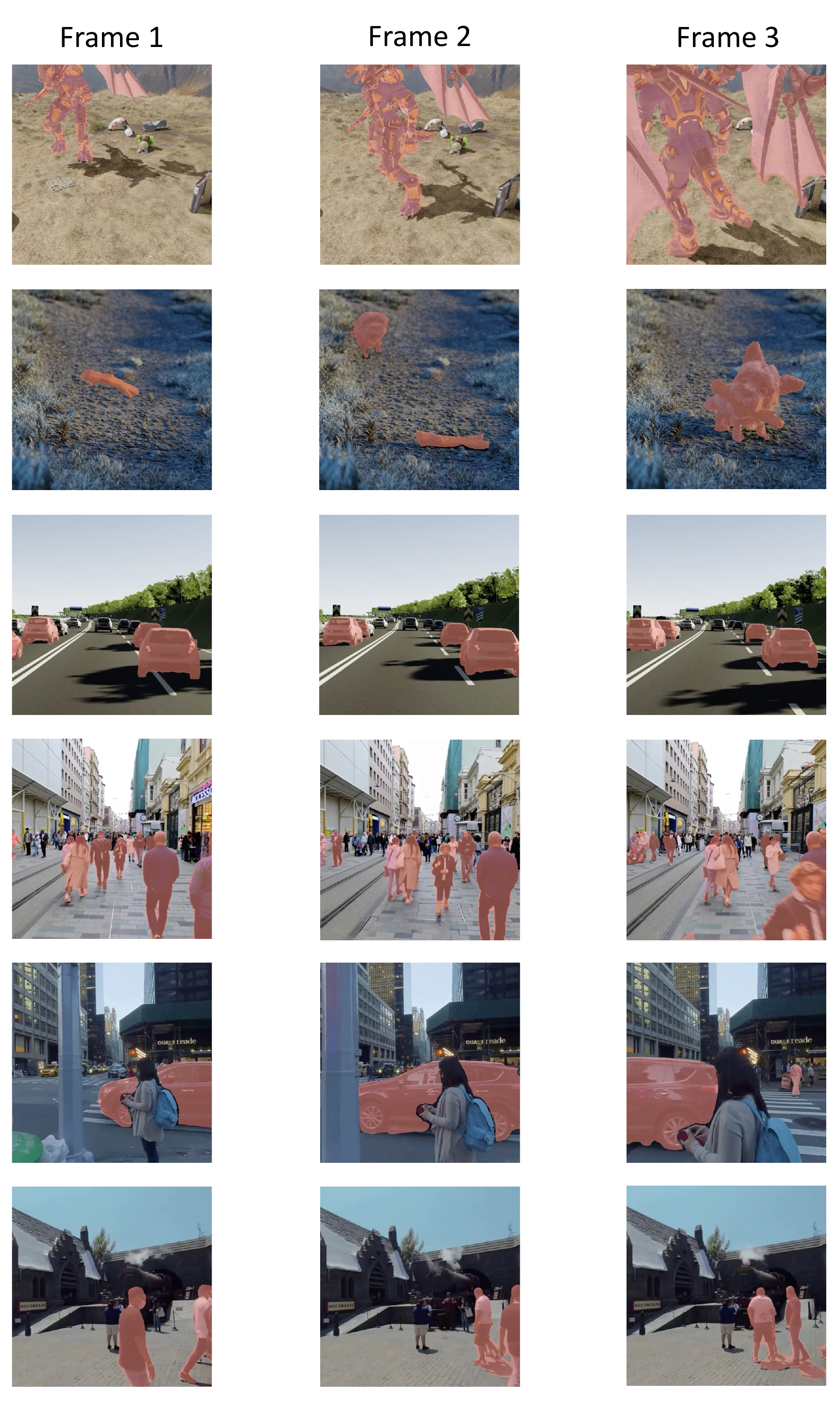}
    \end{center}
    \vspace{-0.2cm}
    \caption{\textbf{Qualitative results for moving object segmentation}. Masks for moving parts, which are filtered from the values of estimated motion maps, are highlighted across frames in light red. Regions with high motion values are regarded as moving parts.}
    \label{fig:seg}
\end{figure*}

We provide more visualization results on novel view synthesis, 3D point tracking, scene flow estimation, and dynamic object segmentation in Figure~\ref{fig:nvs}, \ref{fig:track}, \ref{fig:flow} and \ref{fig:seg} respectively.
Qualitative visualizations of novel view synthesis and 3D point tracking are conducted on the DAVIS dataset~\citep{perazzi2016davis} and TAPVid-3D~\citep{koppula2024tapvid} respectively, which are not included in the training datasets.
Other visualization results are conducted on the test set of the curated datasets to evaluate the generalizability of \name.
XYZ values in the 3D space of motion maps are normalized to $[0,1]$ and treated as RGB channels for visualization. 

\end{document}